\documentclass[10pt,twocolumn]{article}

\usepackage[margin=0.85in,top=1.0in,bottom=1.0in]{geometry}
\usepackage{times}
\usepackage{graphicx}
\usepackage{amsmath,amssymb}
\usepackage{booktabs}
\usepackage{hyperref}
\usepackage{xcolor}
\usepackage{microtype}
\usepackage{caption}
\usepackage{subcaption}
\usepackage{enumitem}
\usepackage{algorithm}
\usepackage{algorithmic}
\usepackage{balance}
\usepackage{natbib}
\usepackage[utf8]{inputenc}
\usepackage[T1]{fontenc}
\usepackage{float} 
\hypersetup{
  colorlinks=true,
  linkcolor=blue!70!black,
  citecolor=blue!70!black,
  urlcolor=blue!70!black
}

\title{%
  \textbf{Real-Time Whole-Body Teleoperation of a Humanoid Robot\\
  Using IMU-Based Motion Capture with Sim2Sim and Sim2Real Validation}
}

\author{%
  Hamza Ahmed Durrani\\
  WeGo Robotics\\
  \texttt{hamzadurrani30@gmail.com}
  \and
  Suleman Khan\\
  WeGo Robotics\\
  \texttt{sulemank@wego-robotics.com}
}
\date{May 2026}

\begin{document}
\maketitle

\begin{figure*}[t]
  \centering
  \includegraphics[width=\textwidth]{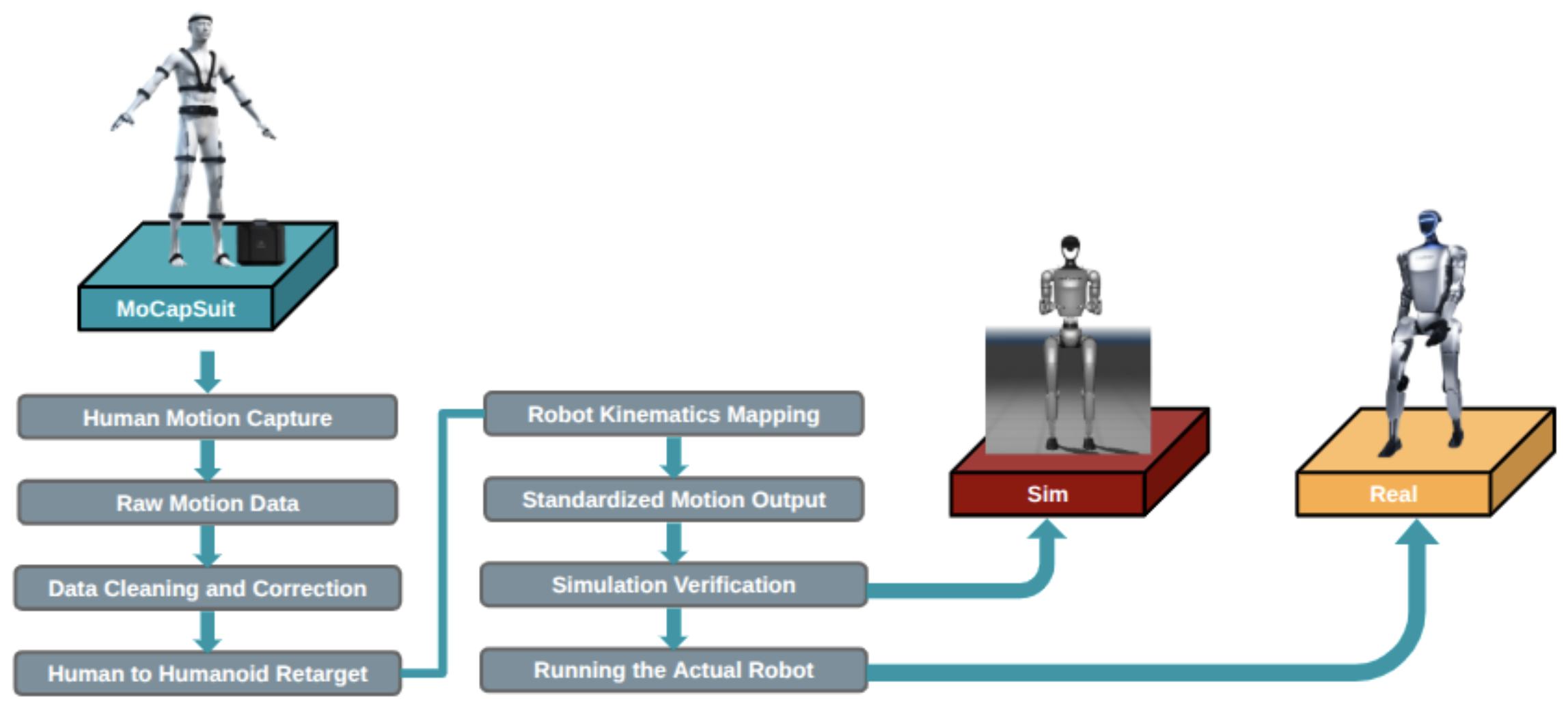}
  \caption{End-to-end teleoperation pipeline: from raw IMU-based motion capture through
           kinematic retargeting to execution on both the MuJoCo simulator and the
           physical Unitree~G1 humanoid robot.}
  \label{fig:pipeline}
\end{figure*}

\begin{abstract}
Stable, low-latency whole-body teleoperation of humanoid robots is an open research
challenge, complicated by kinematic mismatches between human and robot morphologies,
accumulated inertial sensor noise, non-trivial control latency, and persistent
sim-to-real transfer gaps.
This paper presents a complete real-time whole-body teleoperation system that maps human
motion, recorded with a Virdyn IMU-based full-body motion capture suit, directly onto a
Unitree~G1 humanoid robot.
We introduce a custom motion-processing, kinematic retargeting, and control pipeline
engineered for continuous, imperceptibly-low-latency operation without any offline
buffering or learning-based components.
The system is first validated in simulation using the MuJoCo physics model of the
Unitree~G1 (sim2sim), and then deployed without modification on the physical platform
(sim2real).
Experimental results demonstrate stable, synchronized reproduction of a broad motion
repertoire, including walking, standing, sitting, turning, bowing, and coordinated
expressive full-body gestures.
This work establishes a practical, scalable framework for whole-body humanoid
teleoperation using commodity wearable motion capture hardware.

\end{abstract}

\section{Introduction}
\label{sec:intro}

Humanoid robots are increasingly expected to operate in unstructured human environments,
performing manipulation, locomotion, and expressive interaction tasks that demand rich,
coordinated whole-body motion.
Despite significant advances in actuator technology, sensing, and mechanical design, a
central question remains: how can a humanoid robot generate responsive, stable, and
naturalistic full-body movements with minimal engineering effort?

Classical approaches to this problem fall into three broad categories: pre-scripted
motion playback~\cite{unitree_g1}, offline trajectory
optimization~\cite{cheng2024expressive}, and learning-based whole-body
controllers~\cite{nvidia_wbc, he2024omnih2o}.
Although these methods have demonstrated impressive results in controlled settings, they
typically require substantial tuning, large demonstration datasets, or lengthy training
procedures before deployment.

Captured human motion offers a compelling alternative: human locomotion and manipulation
are naturally stable, expressive, and data-rich.
Retargeting this motion onto a humanoid in real time enables direct, intuitive
teleoperation without the overhead of model training.
Yet real-time humanoid teleoperation from wearable inertial measurement unit (IMU) suits
remains challenging, primarily due to (i)~kinematic discrepancies between human and
robot link lengths and joint structures, (ii)~high-frequency noise inherent in
IMU-based pose estimation, (iii)~the need for joint-limit-safe motion at robot control
rates, and (iv)~the risk of instability when transitioning from simulation to physical
hardware.

This paper addresses all four challenges and makes the following contributions:

\begin{enumerate}[leftmargin=*, itemsep=2pt, topsep=2pt]
  \item A lightweight, real-time teleoperation pipeline for full-body humanoid control
        that requires no learning and no offline preprocessing.
  \item A unified kinematic retargeting algorithm that operates identically in
        simulation and on physical hardware, enabling a zero-modification sim2real
        transfer.
  \item Systematic staged validation: sim2sim verification in MuJoCo followed by direct
        deployment on a physical Unitree~G1.
  \item Empirical demonstration of diverse whole-body motions including walking, sitting,
        turning, bowing, and coordinated gestures, reproduced with no perceptible
        latency.
\end{enumerate}

The remainder of the paper is organized as follows.
Section~\ref{sec:related} surveys related work.
Section~\ref{sec:mocap} describes the motion capture hardware.
Section~\ref{sec:retarget} details the retargeting and control pipeline.
Sections~\ref{sec:sim2sim} and~\ref{sec:sim2real} present the simulation and real-robot
experiments, respectively.
Section~\ref{sec:discussion} discusses implications and future directions.

\section{Related Work}
\label{sec:related}

\paragraph{Whole-body control for humanoids.}
Hierarchical whole-body controllers that enforce task priorities while respecting joint
limits are well established~\cite{nvidia_wbc}.
Recent neural whole-body controllers, such as those built on reinforcement learning,
have demonstrated robust locomotion and manipulation on physical
platforms~\cite{he2024omnih2o, cheng2024expressive}, but require extensive simulation
training and careful reward engineering.

\paragraph{Motion retargeting and imitation.}
Retargeting human motion onto morphologically different robots has been studied for
animation and robotics alike.
\cite{cheng2024expressive} learn a retargeting policy from video demonstrations to
generate expressive humanoid motion.
\cite{he2024omnih2o} propose a universal dexterous whole-body teleoperation framework
(OmniH2O) that retargets human keypoints onto a humanoid using a trained
controller.
\cite{twist2} extend similar ideas to diverse manipulation scenarios.
In contrast, our approach uses purely kinematic retargeting without any learned
components, making it immediately deployable on new platforms.

\paragraph{IMU-based motion capture.}
IMU suits provide markerless, room-scale-free body tracking suitable for uncontrolled
environments.
Prior work has used such suits for motion dataset collection~\cite{cheng2024expressive}
and offline retargeting, but seldom for continuous real-time humanoid teleoperation with
direct hardware deployment.

\paragraph{Sim-to-real transfer.}
Closing the sim-to-real gap for humanoids is an active research area.
Domain randomization and careful physics calibration are common mitigation
strategies~\cite{nvidia_wbc, he2024omnih2o}.
Our work sidesteps many of these issues by using a kinematic (rather than dynamic)
retargeting layer, which is inherently physics-agnostic and therefore transfers directly
without additional tuning.

\section{Human Motion Capture}
\label{sec:mocap}

Human motion is recorded using a Virdyn full-body IMU motion capture suit equipped with
inertial sensors distributed across all major body segments.
The suit estimates segment orientations and relative joint angles at runtime without
relying on external cameras, optical markers, or structured environments, making it
suitable for unconstrained real-world operation.

\begin{figure}[t]
  \centering
  \includegraphics[width=0.90\columnwidth]{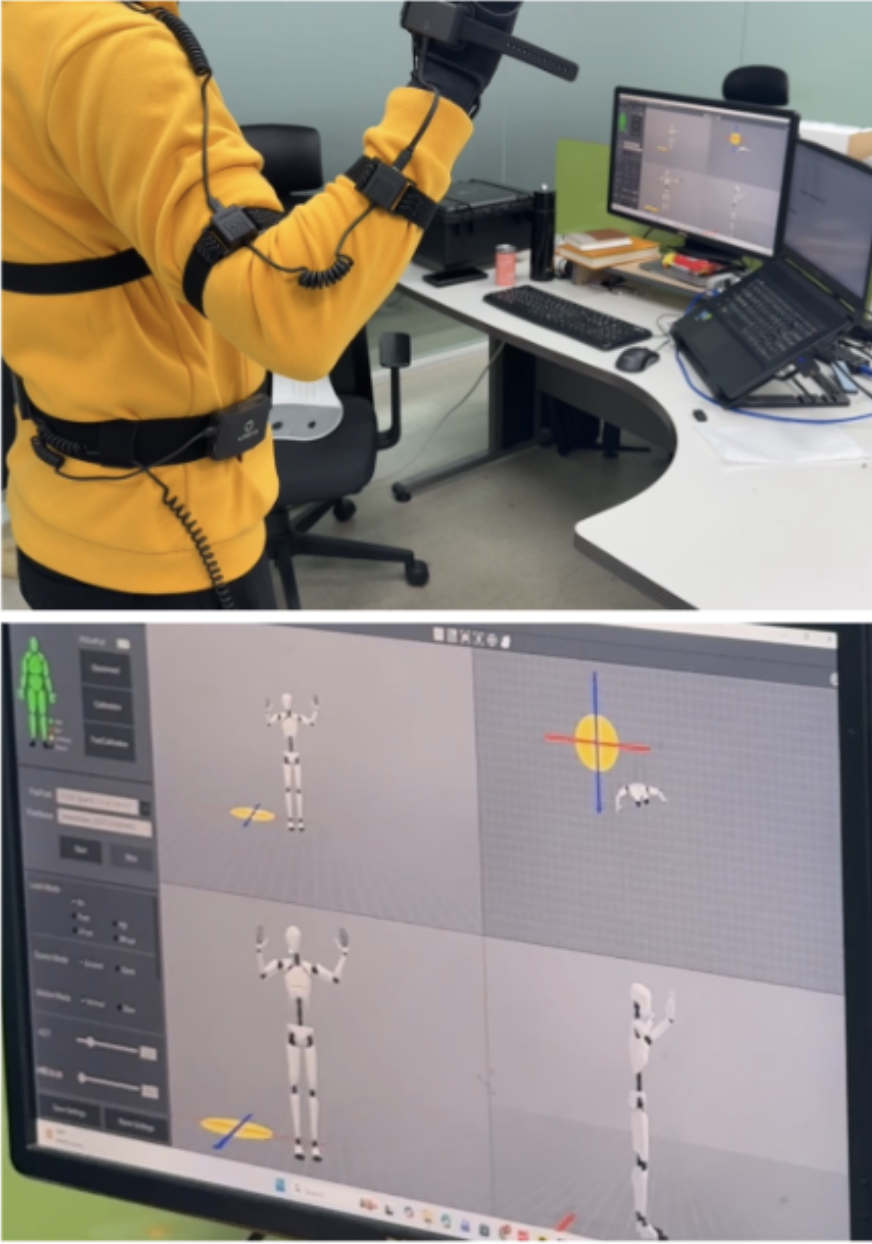}
  \caption{Virdyn IMU suit output visualized in the vendor reference platform, confirming
           full-body skeleton tracking prior to integration with the robot pipeline.}
  \label{fig:mocap}
\end{figure}

Prior to integration with the robot pipeline, the motion capture system was validated
independently using the Virdyn reference visualization platform.
This step confirmed correct sensor calibration, full-body tracking consistency, and
accurate reproduction of diverse upper- and lower-body motions (Fig.~\ref{fig:mocap}).
Only after passing this baseline validation was the suit's output fed into the
retargeting pipeline.

\section{Real-Time Retargeting and Control Pipeline}
\label{sec:retarget}

\subsection{Design Rationale}

A direct, one-to-one mapping of human joint angles to robot joint commands is
infeasible because humans and the Unitree~G1 differ substantially in link proportions,
degrees of freedom, and joint range limits.
Our retargeting algorithm bridges this gap through four tightly coupled modules, all
executing synchronously within the robot control loop.

\subsection{Kinematic Mapping}

Human skeleton joints are mapped to their closest functional counterparts on the
Unitree~G1 kinematic tree.
Where structural differences prevent a direct correspondence (e.g., the human hip
complex versus the robot's three-DoF hip joint), we compute equivalent angles via
geometric projection that preserve the intended motion intent while remaining within the
robot's physical range.

\subsection{Joint Limit Enforcement}

Every mapped joint command is clipped to the robot's hardware joint limits before
transmission.
Soft limits slightly inside the hard mechanical stops are applied to prevent actuator
saturation and reduce wear, particularly for high-velocity expressive motions.

\begin{figure}[H]
  \centering
  \includegraphics[width=0.90\columnwidth]{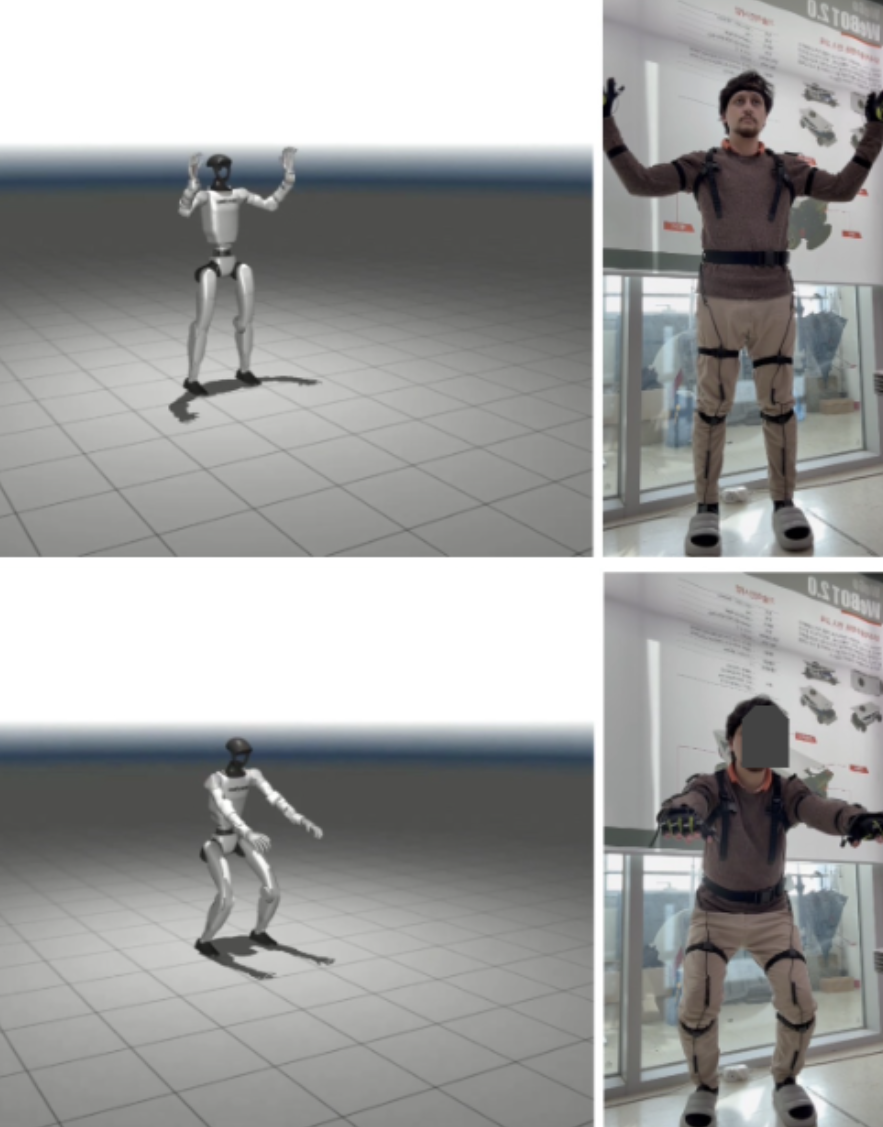}
  \caption{Side-by-side comparison of the human operator and the retargeted MuJoCo
           Unitree~G1 simulation during a representative whole-body motion sequence.}
  \label{fig:sim2sim}
\end{figure}

\subsection{Real-Time Smoothing}

Raw IMU estimates contain high-frequency noise that, if left unfiltered, manifests as
rapid, high-amplitude joint oscillations incompatible with stable locomotion.
We apply a lightweight exponential moving average (EMA) filter per joint, with a time
constant tuned to balance noise attenuation against motion responsiveness.
This choice avoids the computational overhead of more complex filters while achieving
adequate noise rejection at the required control rate.

\subsection{Synchronization}

Upper-body, lower-body, and torso motions are synchronized within a single retargeting
step so that the robot's center-of-mass trajectory remains consistent with the
operator's posture.
The entire pipeline operates in a tight loop with no buffering or batch processing,
guaranteeing that each sensor frame produces a corresponding robot command within the
same control cycle.

A key design advantage of this pipeline is its model-free nature: no neural network
weights, training data, or environment-specific parameters are required.
This makes the system immediately portable to other humanoid platforms with comparable
kinematics.

\section{Simulation Verification (Sim2Sim)}
\label{sec:sim2sim}

Before deploying on physical hardware, the full pipeline was evaluated in the MuJoCo
physics simulator using the official Unitree~G1 model (Fig.~\ref{fig:sim2sim}).
The IMU suit streamed live joint estimates into the retargeting module, whose output was
applied directly to the simulated robot.
This stage allowed us to verify retargeting correctness, identify pathological joint
configurations, and confirm motion stability across the full range of tested behaviors.

The simulation stage served two explicit purposes.
First, it validated that the retargeting algorithm produces physically plausible
configurations with no joint limit violations, no self-collisions, and no abrupt
discontinuities in joint velocity.
Second, because the same retargeting code runs unmodified in both simulation and on the
real robot, passing sim2sim verification constitutes strong evidence that the pipeline
will transfer successfully to hardware.

\section{Real Robot Deployment (Sim2Real)}
\label{sec:sim2real}

Following successful sim2sim validation, the identical pipeline was deployed on the
physical Unitree~G1 humanoid without any modification to the motion processing or
retargeting logic.
The transition required no additional domain adaptation, parameter retuning, or
controller retraining, which underscores the value of a physics-agnostic kinematic
retargeting design.

\begin{figure}[t]
  \centering
  \includegraphics[width=0.85\columnwidth]{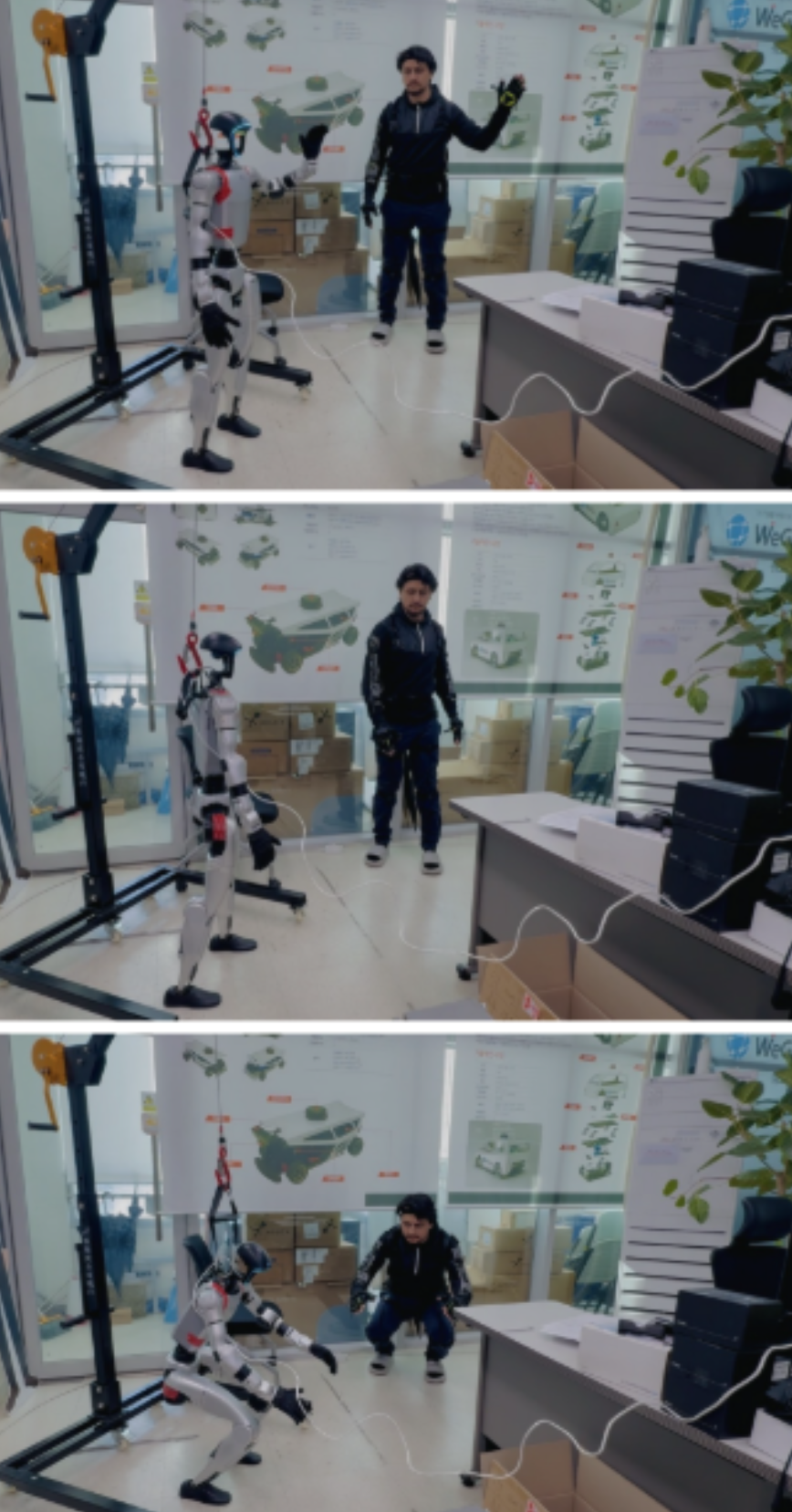}
  \caption{Synchronized human to robot teleoperation on the physical Unitree~G1.
           The operator (left) performs walking and sitting motions that are reproduced
           in real time by the robot (right) with no perceptible latency.}
  \label{fig:sim2real}
\end{figure}

The robot demonstrated real-time reproduction of the following motion categories
(Fig.~\ref{fig:sim2real}):

\begin{enumerate}[leftmargin=*, itemsep=2pt, topsep=2pt]
  \item \textbf{Locomotion:} forward and backward walking, stepping in place.
  \item \textbf{Postural transitions:} sitting down and standing up from a chair.
  \item \textbf{Orientation changes:} in-place turning and lateral weight shifts.
  \item \textbf{Expressive gestures:} bowing, waving, and coordinated arm motions.
  \item \textbf{Compound motions:} simultaneous arm gestures during locomotion.
\end{enumerate}

No perceptible delay was observed between the operator's motion and the robot's
execution, confirming that the pipeline meets real-time performance requirements.
Qualitatively, the motion reproductions were smooth and stable throughout all tested
sequences, with the robot maintaining balance without any additional stabilization
controller beyond the Unitree~G1's onboard low-level joint servo.

\section{Discussion and Impact}
\label{sec:discussion}

\subsection{Summary of Findings}

This work demonstrates that IMU-based full-body teleoperation of a humanoid robot is
practically achievable with a purely kinematic, learning-free pipeline.
The proposed system is stable, low-latency, and transfers directly from simulation to
real hardware without modification, a combination that has been difficult to achieve with
learning-based approaches due to their inherent dependence on training distributions and
simulation fidelity.

\subsection{Advantages over Prior Approaches}

Compared to reinforcement-learning-based whole-body controllers such as those of
\cite{nvidia_wbc} and \cite{he2024omnih2o}, our system requires no training data,
no simulation-to-real domain randomization, and no reward engineering.
Compared to offline retargeting pipelines, it supports true real-time operation.
The absence of learned components also means that the system's behavior is fully
interpretable and deterministic, which is advantageous in safety-critical deployment
scenarios.

\subsection{Limitations and Future Work}

The current system relies on the Unitree~G1's onboard servo controllers for low-level
stability.
In highly dynamic motions such as rapid direction changes or large-amplitude arm swings at
speed, the robot's balance could be improved by incorporating an online whole-body
momentum controller or a model-predictive footstep planner.
Additionally, the EMA filter introduces a small, motion-speed-dependent phase lag, and an
adaptive Kalman filter could further reduce this effect.

Future directions include: (i)~automatic retargeting parameter adaptation for other
humanoid platforms, (ii)~integration with imitation learning pipelines to bootstrap
neural whole-body controllers from teleoperated demonstrations~\cite{cheng2024expressive},
and (iii)~extension to dexterous hand manipulation using finger-level IMU data or
vision-based hand tracking.

\bibliographystyle{plain}

\end{document}